\documentclass[10pt,twocolumn,letterpaper]{article}

\usepackage{iccv}
\usepackage{times}
\usepackage{epsfig}
\usepackage{graphicx}
\usepackage{amsmath}
\usepackage{amssymb}
\usepackage{microtype}

\usepackage{caption}
\usepackage{subcaption}

\usepackage{array}
\usepackage{booktabs}
\usepackage{tabularray}
\usepackage{xspace}
\usepackage{nicefrac}

\usepackage{siunitx}
\usepackage{xspace}
\usepackage{multirow}
\usepackage[accsupp]{axessibility}  

\usepackage[table]{xcolor}

\definecolor{best}{RGB}{255, 220, 200}
\definecolor{second}{RGB}{255, 255, 200}

\usepackage[pagebackref=true,breaklinks=true,letterpaper=true,colorlinks,bookmarks=false]{hyperref}

\usepackage[capitalize]{cleveref}
\crefname{section}{sec.}{secs.}
\Crefname{section}{Sec.}{Secs.}
\crefname{table}{tab.}{tabs.}
\Crefname{table}{Tab.}{Tabs.}
\crefname{figure}{fig.}{figs.}
\Crefname{figure}{Fig.}{Figs.}
\crefname{equation}{eq.}{eqs.}
\Crefname{equation}{Eq.}{Eqs.}

\newcommand{\Ak}[1]{\textcolor{black}{#1}}
\newcommand{\ps}[1]

\iccvfinalcopy 

\newcommand{\thename}{DarSwin\xspace}

\ificcvfinal\fi

\begin{document}

\title{\thename: Distortion Aware Radial Swin Transformer}

\author{Akshaya Athwale{$^1$}, Arman Afrasiyabi{$^{3}$}, Justin Lagüe{$^{1}$}, Ichrak Shili{$^{1}$},\\ Ola Ahmad{$^{1,2}$}, Jean-François Lalonde{$^{1}$} \\
{\small{$^{1}$}Université Laval \quad {$^{2}$}Thales Digital Solutions \quad {$^{3}$} Yale University}
}

\maketitle
\ificcvfinal\thispagestyle{empty}\fi

\begin{abstract}
Wide-angle lenses are commonly used in perception tasks requiring a large field of view. Unfortunately, these lenses produce significant distortions making conventional models that ignore the distortion effects unable to adapt to wide-angle images. In this paper, we present a novel transformer-based model that automatically adapts to the distortion produced by wide-angle lenses. We leverage the physical characteristics of such lenses, which are analytically defined by the radial distortion profile (assumed to be known), to develop a distortion aware radial swin transformer (DarSwin). In contrast to conventional transformer-based architectures, DarSwin comprises a radial patch partitioning, a distortion-based sampling technique for creating token embeddings, and an angular position encoding for radial patch merging.
We validate our method on classification tasks using synthetically distorted ImageNet data and show through extensive experiments that DarSwin can perform zero-shot adaptation to unseen distortions of different wide-angle lenses. Compared to other baselines, DarSwin achieves the best results (in terms of Top-1 accuracy) with significant gains when trained on bounded levels of distortions (very-low, low, medium, and high) and tested on all including out-of-distribution distortions. 
%
The code and models are publicly available at \href{https://lvsn.github.io/darswin/}{https://lvsn.github.io/darswin/} 
\end{abstract}
\section{Introduction}


Wide field of view (FOV) lenses are becoming increasingly popular because their increased FOV minimizes cost, energy, and computation since fewer cameras are needed to image the entire environment.
They are having a positive impact on many applications, including security ~\cite{kim2015fisheye}, augmented reality (AR)~\cite{schmalstieg2017ar}, healthcare and more particularly, autonomous vehicles \cite{deng2019restricted,yogamani2019woodscapes}, which require sensing their surrounding $360^{\circ}$ environment.
\begin{figure}[t] 
    \centering
    \footnotesize
    \setlength{\tabcolsep}{1pt}
    \begin{tabular}{cccc} 
    \includegraphics[width=0.99\linewidth, angle=0]{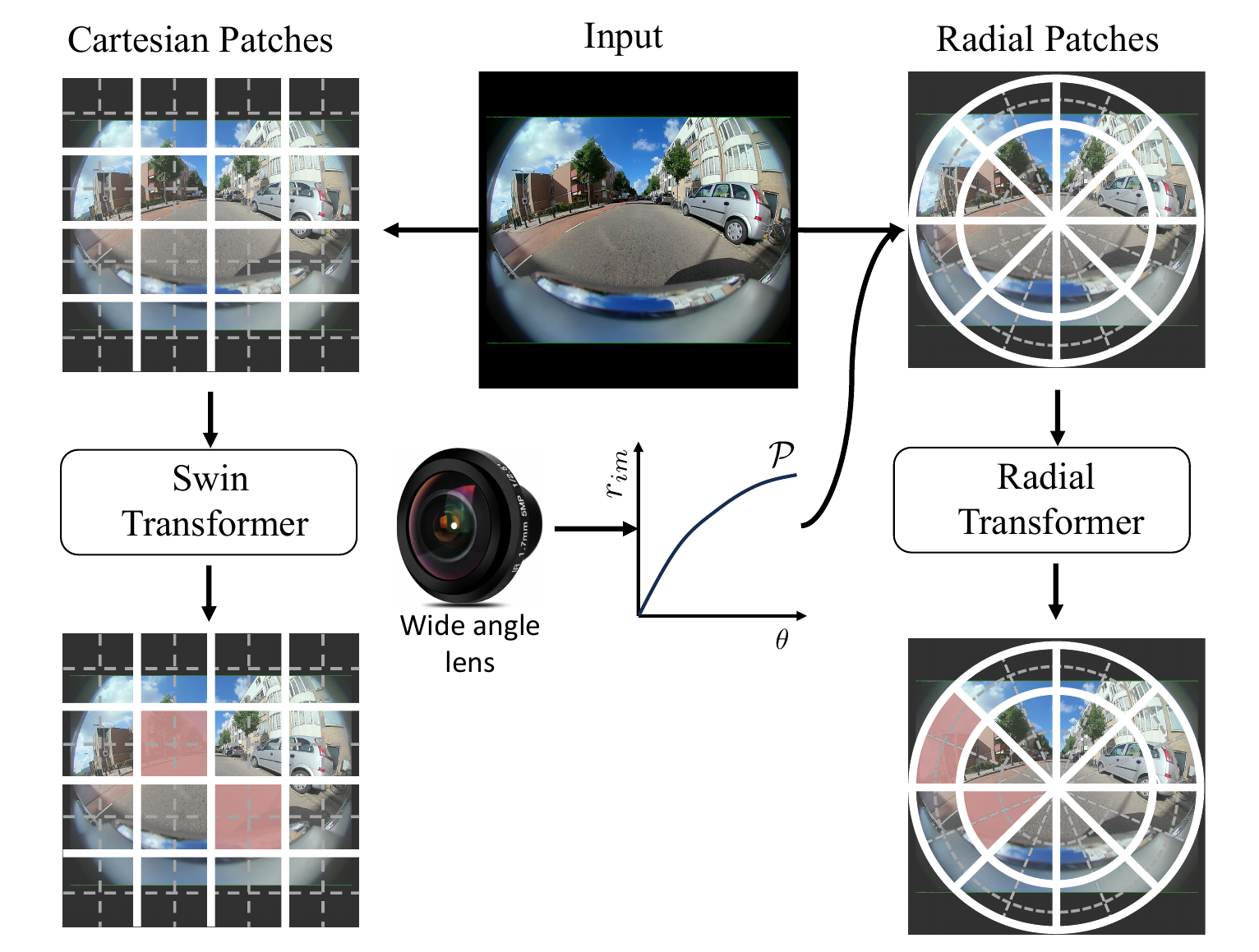} \\
    \ \ \   \rotatebox{0}{\footnotesize{ \quad Swin~\cite{liu2021swin}} \quad \ \  \ \ \ \  \quad  \quad  \quad  \quad    \quad  \quad  \quad   \quad  \quad   \quad \quad \      \footnotesize{DarSwin} (ours)  }   \\     
    \end{tabular}
    \caption[]{Illustration of (cartesian) Swin~\cite{liu2021swin} (left) and our (radial) \thename (right) given a wide-angle image (middle). While Swin~\cite{liu2021swin} computes attention on the predefined windows over square image patches (bottom left orange region), \thename performs radial transformations using distortion-aware radial patches and computes the attention on windows defined over radial patches (shown in orange region bottom right), which enables greater generalization capabilities across different lenses.}
    \label{fig:teaser}
\end{figure}

Unfortunately, such wide angle lenses create significant distortions in the image since the perspective projection model no longer applies: straight lines appear curved, and the appearance of the same object changes as a function of its position on the image plane. This distortion breaks the translation equivariance assumption implicit in convolution neural networks (CNNs) and therefore limits their applicability. This problem is further exacerbated by the diversity in lens distortion profiles: as we will demonstrate, a network naively trained for a specific lens tends to overfit to that specific distortion, and does not generalize well when tested on another lens. Just as methods are needed to address the ``domain gap''~\cite{huang2018multimodal} from dataset bias~\cite{torralba2011unbiased,khosla2012undoing}, we believe we must also bridge the ``distortion gap'' to truly make wide angle imaging applicable.

One popular strategy to bridge the distortion gap is to cancel the effect of distortion on the image plane by warping the input image back to a perspective projection model according to calibrated lens parameters. Conventional approaches can then be trained and tested on the resulting ``un-distorted'' images. A wide array of such methods, ranging from classical~\cite{brousseau2019calibration,ramalingan2017unifying,zhang2015line,melo2013unsupervised,kannala2006generic} to deep learning~\cite{yin2018fisheyerectnet,xue2019learning}, have been proposed. Unfortunately, warping a very wide angle image to a perspective projection tends to create severely stretched images and restricts the maximum field of view since, in the limit, a point at \ang{90} azimuth projects at infinity. Reducing the maximum field of view defeats the purpose of using a wide angle lens in the first place. Other projections are also possible (e.g., cylindrical~\cite{plaut2021fisheye}) but these also tend to create unwanted distortions.

Recently, methods that break free from the ``undistort first'' strategy aim to reason directly about the images without undistorting them. For example, methods like \cite{playout2021adaptable,ahmad2022fisheyehdk} use deformable convolutions~\cite{dai2017deformable,zhu2019deformable} to adapt convolution kernels to the lens distortion. However, the high computational cost of deformable CNNs constrains the kernel adaptation to a few layers inside the network. Other approaches like spherical CNNs~\cite{cohen2018spherical} or gauge equivariant CNN~\cite{cohen2019gauge} can adapt to different manifolds but their applicability for lens distortion has not been demonstrated. Finally, vision transformers~\cite{dosovitskiy2020vit} and their more recent variants~\cite{zhou2021deepvit,liu2021swin,xia2022vision} could also better bridge the ``distortion gap'' since they do not assume any prior structure other than permutation equivariance, but their cartesian partitioning of the image plane do not take lens geometry into account (see \cref{fig:teaser}).

In this paper, we present \thename, a transformer-based architecture that adapts its structure to the lens distortion profile, which is assumed to be known (i.e. the camera is calibrated). Our method, inspired by the recent Swin transformer architecture~\cite{liu2021swin}, leverages a distortion aware sampling scheme for creating token embeddings, employs polar patch partitioning and merging strategies, and relies on angular relative positional encoding. 
This explicitly embeds knowledge of the lens distortion in the architecture and makes it much more robust to the ``distortion gap'' created by training and testing on different lenses. 


The main contribution of this paper is a novel transformer-based encoder that automatically adapts to the (known) lens distortion profile, which relies on a combination of the following novel distortion aware components: polar patch partitioning, distortion-based sampling scheme for creating token embeddings along with a jittering technique for better generalization, and angular relative position encoding for radial patch merging.
We show, through extensive classification experiments, that \thename can perform zero-shot adaptation (without pretraining) across different lenses. Indeed, when our method is trained on a restricted set of distortions, we observe that it is much more robust to changes in distortions at test time than all of the compared alternatives, including baseline Swin~\cite{liu2021swin} (applied on both distorted and undistorted images) and deformable attention transformer ~\cite{xia2022vision}. 

\section{Related work}


\paragraph{Panoramic distortion} Panoramic images span the full \ang{360} field of view and are most commonly projected onto a plane using an equirectangular projection creates severe distortions especially near the poles. Many works have explicitly designed approaches to deal with equirectangular distortion, including depth estimation~\cite{zioulis2018omnidepth}, saliency detection~\cite{yun2022panoramic}, segmentation~\cite{zhang2022bending}, layout estimation~\cite{fernandez2018layouts}, and object detection~\cite{su2019kernel} to name a few. However, as with \cite{cohen2018spherical}, these are specifically designed for spherical distortion and do not generalize to wide-angle lenses.

\paragraph{Image undistortion}
Applying tasks like classification and object recognition on wide-angle images is relatively recent~\cite{Rashed_2021_WACV,inproceedings,plaut2021fisheye, Ye2020UniversalSS, yogamani2019woodscapes} due to the presence of distortions in the image. In computer vision, the application of wide-angle images ranges from visual perception~\cite{kumar2021omnidet} to autonomous vehicle cameras~\cite{yogamani2019woodscapes, kumar2022surround, liao2022kitti}. In this respect, the initial studies mainly focused on correcting the distortion of the image \cite{xue2019learning, article1, yin2018fisheyerectnet, zhang2015line, 9980359}. Recently, many convolutional-based and attention-based models have been proposed that try to directly reason on wide-angle images without relying on distortion correction models. 

\paragraph{Convolution-based approaches} 
CNNs~\cite{krizhevsky2012classif, VGG, googlenet} are particularly well-suited for perspective images due to their implicit bias and translational equivariance \cite{GDL}. Methods like \cite{8500456, inproceedings, Rashed_2021_WACV} tries to adapt CNNs on fisheye images for tasks such as object detection. However, the distortion caused by wide angle images breaks this symmetry, which reduces the generalization performance of CNNs. Deformable convolutions~\cite{dai2017deformable} (later extended in \cite{zhu2019deformable}) learn a deformation to be applied to convolution kernels which can provide greater flexibility at the cost of significant additional computation. Closer to our work, \cite{ahmad2022fisheyehdk, playout2021adaptable,deng2019restricted} use such deformable CNNs to understand the distortion in a fisheye image. In contrast, our \thename leverages attention-based mechanisms rather than convolution. Recently, Jang et al.~\cite{jang2022dada} proposed a framework for distortion-aware domain adaptation, where a generator network is trained to transform an image to a different lens profile. In contrast, our method can adapt to a different lens without additional training. 



\paragraph{Self-attention-based approaches} Vision transformers (ViT)~\cite{dosovitskiy2020vit} use self-attention mechanisms~\cite{vaswani2017attention} computed on image patches rather than performing convolutions. Unlike CNNs, a ViT does not have a fixed geometric structure in its architecture: any extra structure is given via positional encoding. More recently, the Swin transformer architecture~\cite{liu2021swin} proposes a multi-scale strategy of window-based attention. Later, deformable attention transformer (DAT)~\cite{xia2022vision}, adapts the concept of deformable CNNs~\cite{dai2017deformable,zhu2019deformable} to increase adaptability. Unlike existing transformer architectures, \thename explicitly embeds the distortion into its structure using polar sampling, patch partition, window-based self-attention, and angular positional encoding.

\section{Image formation}
\label{sec:background}

\begin{figure*}[!ht]
\centering
\includegraphics[width=\linewidth]{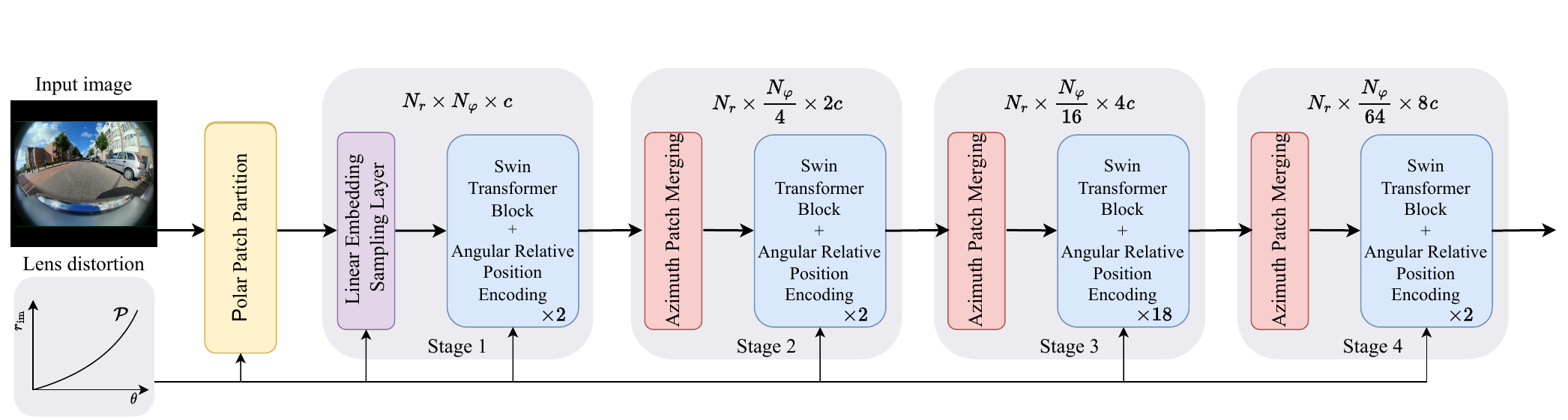}
\caption[]{Overview of our distortion aware transformer architecture, \thename (Azimuth Merge). It employs hierarchical layers of Swin-S transformer blocks~\cite{liu2021swin} interspersed with patch merging layers. To make it adapt to lens distortion, the patch partition, linear embedding, and patch merging layers all take the lens projection curve $\mathcal{P}$ (c.f. \cref{sec:background}) as input.}
\label{fig:darswin-arch}
\end{figure*}

We begin with a brief review of lens distortion models relevant to this work. The pixel coordinates $\mathbf{p}_\mathrm{im}$ of a 3D point $\mathbf{p}_\mathrm{w} = [x, y, z]^\mathsf{T}$ in world coordinates are given by
\begin{equation}
\mathbf{p}_{\mathrm{im}} = [u, \; v]^\mathsf{T} = \mathcal{P}(\mathbf{p}_{\mathrm{w}}) \,,
\label{eq:formation-cartesian}
\end{equation}
where $\mathcal{P}$ is a 3D-to-2D projection operator, including conversion from homogeneous coordinates to 2D. Here, without loss of generality, the camera is assumed to be at the world origin (so its rotation and translation are ignored).

To represent wide-angle images, it is common practice to use a projection model that describes the relationship between the radial distance $r_\mathrm{im} = \sqrt{u^2 + v^2}$ from the image center and the incident angle $\theta = \arctan ( \nicefrac{\sqrt{x^2 + y^2}}{z} )$. This relationship takes the generic form
\begin{equation}
r_{\mathrm{im}} = \mathcal{P}(\theta) \,.
\label{eq:formation-radial}
\end{equation}
Within the scope of our interest, we consider three types of projections.

\paragraph{Perspective projection.}
Under the perspective projection model, the projection operator $\mathcal{P}$ takes the form \cite{kannala2006generic}
\begin{equation}
\mathcal{P}_\mathrm{pers}(\theta) \equiv r_{\mathrm{im}} = f \tan (\theta) \,,
\end{equation}
where $f$ is the focal length (in pixels). The perspective projection is the rectilinear model of pinhole lenses. Wide-angle lenses disobey the law of perspective projection and therefore cause non-linear distortions.   

\paragraph{Polynomial projection.}
In the case of wide-angle lenses, there are several classical projection models \cite{beck1925apparatus,hill1924lens,fleck1995perspective,miyamoto1964fisheye} giving different formulas for $\mathcal{P}$; see \cite{hughes2010accuracy} for a detailed analysis of the accuracy of such models. A unified, more general, model is defined as an $n-$degree polynomial and given by
\begin{equation}
\mathcal{P}_\mathrm{poly}(\theta) \equiv r_{\mathrm{im}} = a_1\theta + a_2\theta^2 + ... + a_n \theta^n \,.
\end{equation}
For example, the WoodScape dataset~\cite{yogamani2019woodscapes} employs a 4-degree ($n=4$) polynomial for their lens calibration. We adopt this polynomial function to define the lens projection curve used in our method (\cref{sec:methodology}).

\paragraph{Spherical projection.}
The spherical projection model~\cite{barreto2006unifying,mei2007single} describes the radial distortion by a \emph{single}, \emph{bounded} parameter $\xi \in [0, 1]$\footnote{$\xi$ can be slightly greater than 1 for certain types of catadioptric cameras~\cite{ying2004consider} but this is ignored here.}. It projects the world point $\mathbf{p}_{\mathrm{w}}$ to the image as follows
%
\begin{equation}
[u, v]^\mathsf{T} = \left[ \frac{xf}{\xi ||\mathbf{p}_\mathrm{w}||+z},\frac{yf}{\xi ||\mathbf{p}_\mathrm{w}||+z} \right]^\mathsf{T} \,.
\label{eqn:sphproj}
\end{equation}
We employ this model in experiments (\cref{sec:experiments}) because of its ability to represent distortion with a single parameter.
%


\section{Methodology}
\label{sec:methodology}

\subsection{Overview}

\Cref{fig:darswin-arch} shows an overview of our proposed distortion aware transformer architecture. It accepts as input a single image and its distortion parameters in the form of a lens projection curve $\mathcal{P}(\theta)$ (see \cref{eq:formation-radial}). The image domain is first segmented into patches according to a polar partitioning module (\cref{sec:polar-partition}). Then, a linear embedding is computed from sampled points (\cref{sec:linear-embedding}), reshaped into a radial-azimuth projection and fed to the first Swin transformer block. The attention mechanism employs an angular relative positional encoding scheme guided by the lens curve. This is followed by three blocks performing patch merging (\cref{sec:patch-merging}) and additional Swin transformer blocks. More details on each of these steps are provided below.

\begin{figure}[t]
  \centering
  \footnotesize
  \includegraphics[width=0.45\textwidth]{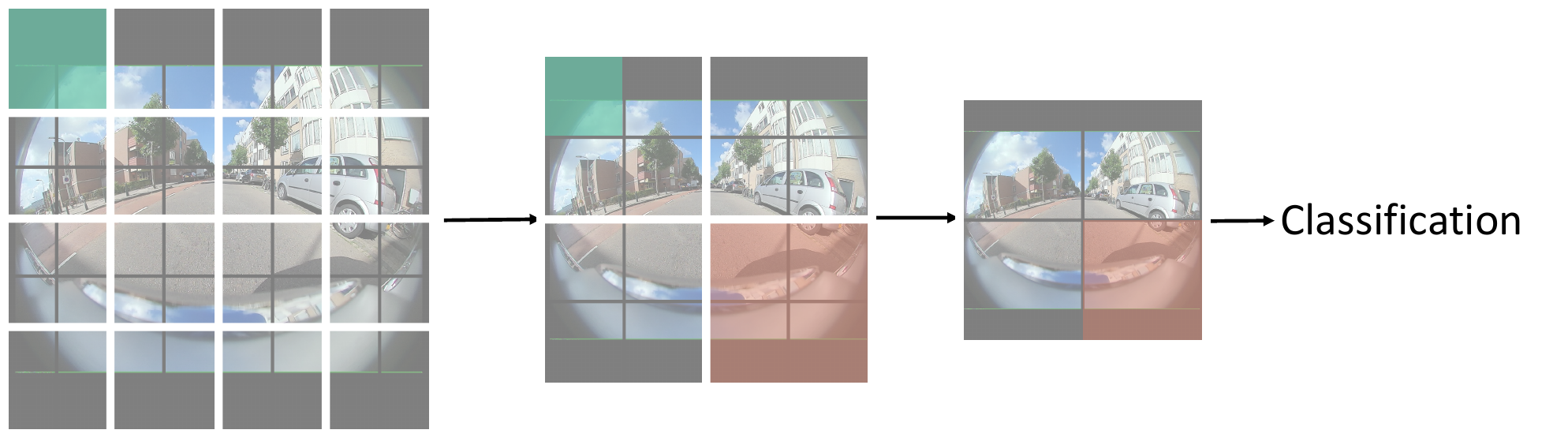} \\
  (a) Swin~\cite{liu2021swin} \\
  \includegraphics[width=0.45\textwidth]{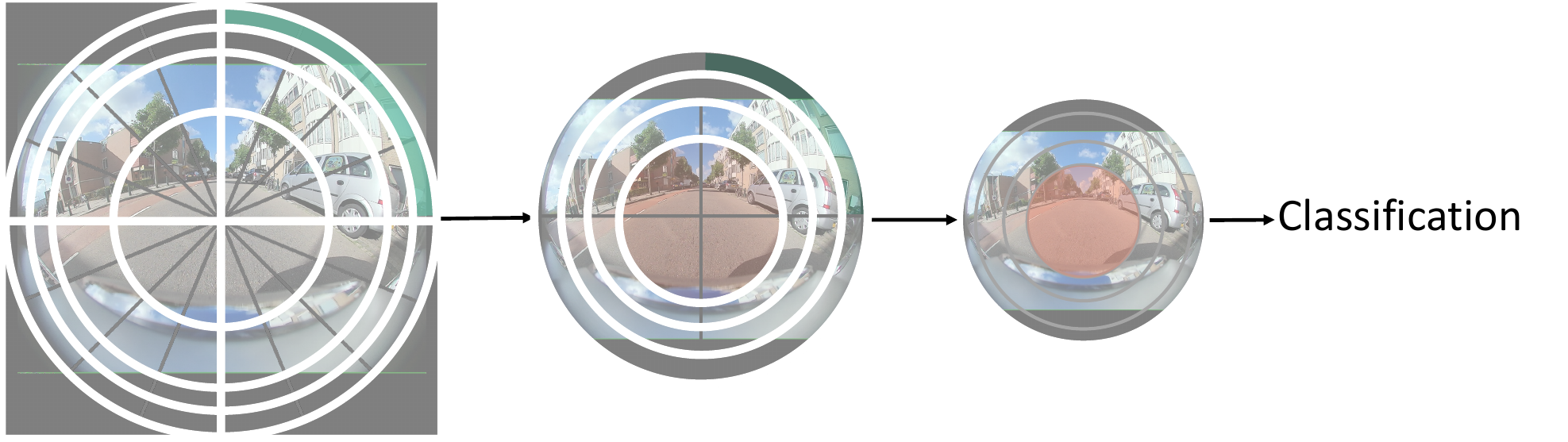} \\
  (b) \thename
  \caption[]{Illustration of the difference between (a) Swin~\cite{liu2021swin}, which uses Cartesian image patches (shown in grey) and windows (shown in white borders) by making divisions along image axes, Swin performs attention on a Cartesian window and merges 2 $\times$ 2 neighbourhood patches to build a hierarchical structure (shown in the green and orange shaded region) (b) our proposed \thename uses polar image patches by making divisions along radius and azimuth. In our case, the images patches (shown in grey) and windows (shown in white borders) are defined along azimuth and merged along azimuth as (shown in the green and orange shaded region)}
  \label{fig:swinvdarswin}
\end{figure}

\subsection{Polar partition}
\label{sec:polar-partition}

\begin{figure}[t]
  \centering
  \includegraphics[width=0.80\linewidth]{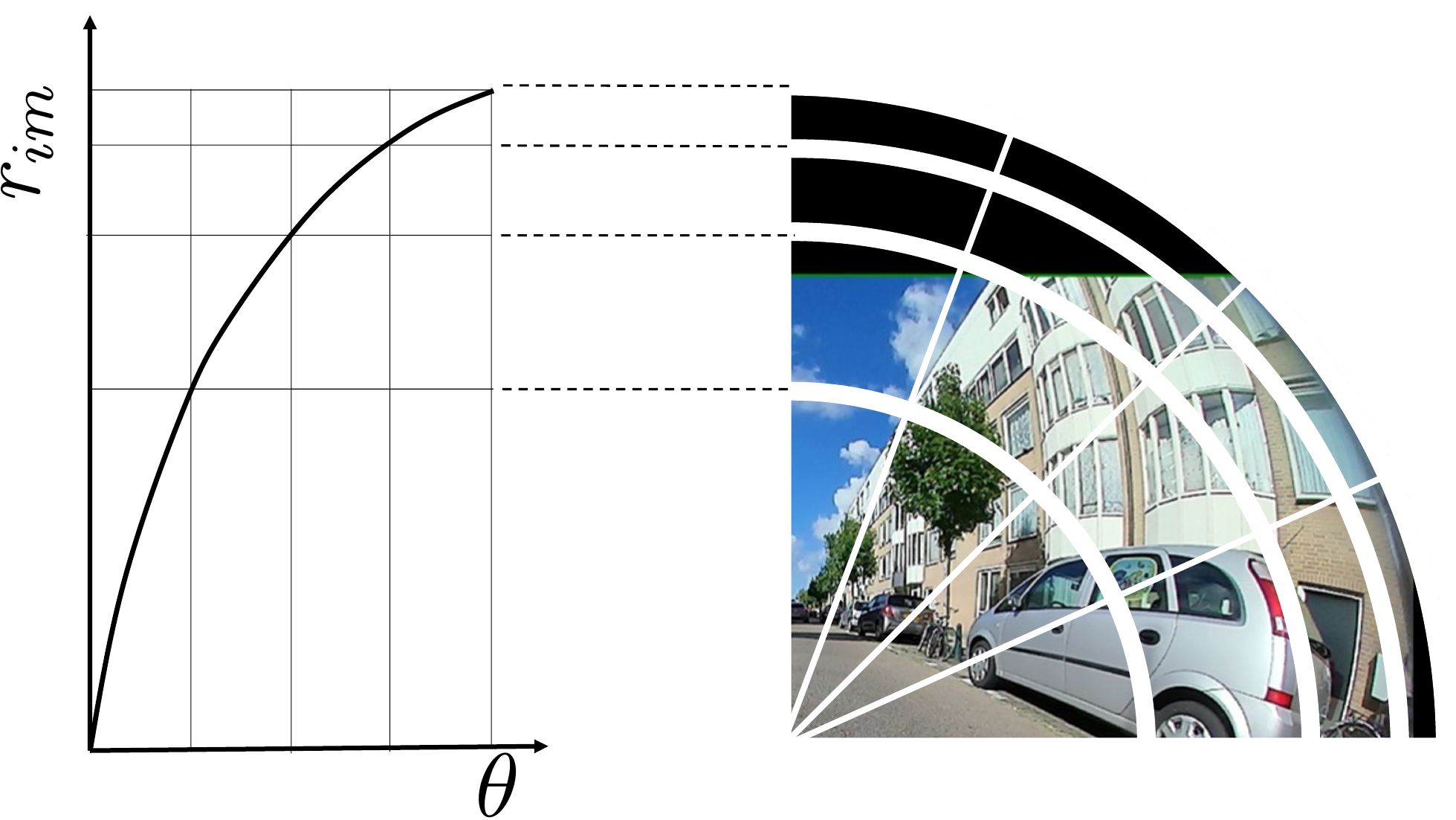}
  \caption{Example of distortion aware polar patch partition. Here, $N_r=4$ and $N_\varphi = 4$ partitions along radius and azimuth respectively are used to illustrate (right, only the top-right quadrant of the image is shown). While the azimuth partitioning is performed in an equiangular fashion, the radial dimension takes the lens distortion curve (left) into account. The field of view along the incident angle $\theta$ is split into $N_r$ equal parts (left) and corresponding radial are obtained from the distortion curve. Hence for different lenses we can have different radii depending on the distortion parameters. }
  \label{fig:patch-partition}
\end{figure}

The first step of our proposed architecture is to partition the image domain (defined by a 2D plane) into patches. As opposed to Swin which performs the split in cartesian coordinates (\cref{fig:swinvdarswin}a), \thename employs a polar patch partitioning strategy (\cref{fig:swinvdarswin}b). After centering a polar coordinate system on the image center (assumed to be known), we first split according to azimuth angle $\varphi$ (in the image plane) in $N_\varphi$ equiangular regions. For the radial dimension, we split the image into $N_r$ radial regions such that the splits are equiangular in $\theta$ and obtain the corresponding radii using the lens projection function $\mathcal{P}(\theta)$ (see \cref{eq:formation-radial,fig:patch-partition}). The total number of patches is therefore $N_\varphi \times N_r$. In our experiments, we set $N_r=16$ and $N_\varphi = 64$ for an input image of size $64\times64$ pixels.

\subsection{Linear embedding}
\label{sec:linear-embedding}

\begin{figure}[!h]
  \centering
  \includegraphics[width=1.0\linewidth]{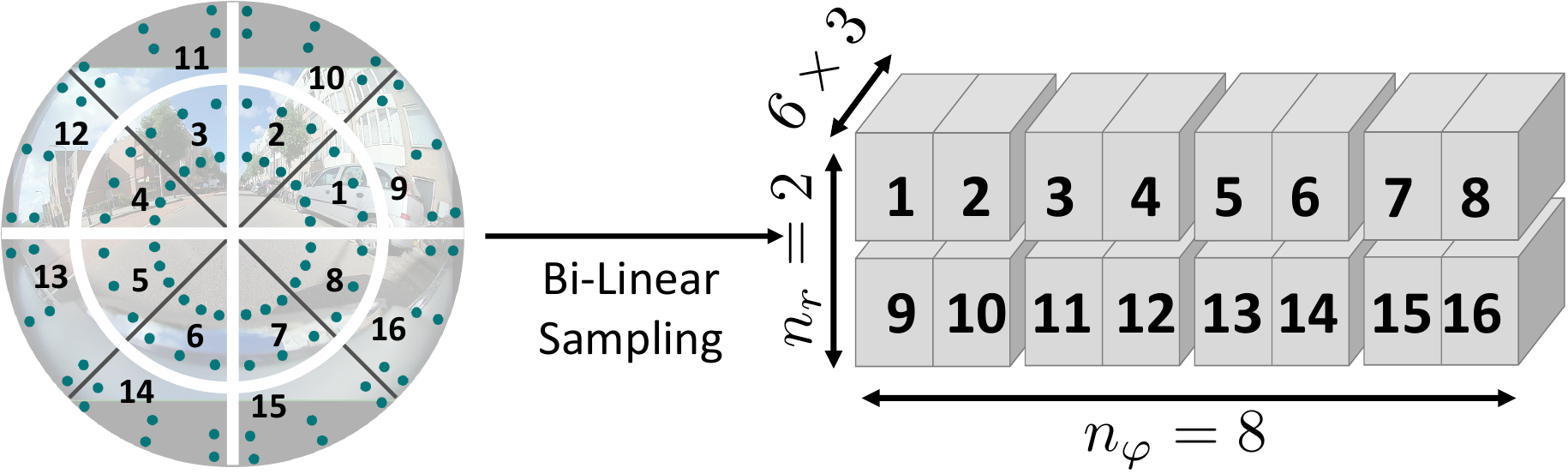}
  \caption{Example of sampling strategy on patch partitions with a total of 16 patches (here, the number of divisions along radius and azimuth are $N_r = 2$ and $N_\varphi = 8$ respectively) with window size $(1, 2)$, i.e. ($M_r = 1, M_\varphi = 2$). We use bilinear interpolation to sample RGB values from the image (6 blue dots per patch) and arrange them in polar coordinates.}
  
  \label{fig:sampling}
\end{figure}

The resulting image patches created by this approach have unequal amount of pixels. Therefore, we rely on a distortion aware sampling strategy to obtain the same number of points for each patch. To sample from the images, we define the number of sampling points along the radius and azimuth, as shown in \cref{fig:sampling}, and adapt the pattern according to each partition. In our experiments, we set 10 sampling points along radius and azimuth for each patch. Points are sampled in an equiangular fashion along the azimuth direction. For the radial dimension, we sample according to the same pattern as the polar partitioning (\cref{sec:polar-partition}); that is, we split in an equiangular fashion according to $\theta$ and obtain the corresponding radii using $\mathcal{P}(\theta)$ (\cref{eq:formation-radial}) as shown in \cref{fig:sampling}. The input image is sampled using bilinear interpolation, and samples are arranged in a polar format as illustrated in \cref{fig:sampling}. The resulting sample values are then fed into a linear embedding layer to produce token embeddings as input for the transformer block.

\subsection{Window-based self attention}
\label{sec:window-attn}

The Swin transformer architecture~\cite{liu2021swin} uses window-based self-attention, where attention is computed on non-overlapping windows of $M\times M$ patches. Here, we maintain that strategy, but the polar nature of our patches allows for an additional design choice. Namely, the number of patches along azimuth $M_\varphi$ and radius $M_r$ can be different. 
As shown in \cref{tab:detailed architecture}, we define two variants: $M$ patches along both azimuth and radius, $M_r = M_\varphi = M$ (DarSwin-RA); or $M_\varphi = M^2$ patches along azimuth and $M_r = 1$ patch along radius (DarSwin-A). Note that we break this rule when the input resolution across radius or azimuth becomes lower than $M$ to maintain the number of patches in each window equals to $M^{2}$ at each Stage (see \cref{tab:detailed architecture}). 
We experimentally found (see \cref{sec:ablation}) that computing attention across $M^2$ patches along the azimuth (DarSwin-A) yielded improved performance. 




Similarly, shifted window self-attention (used in \cite{liu2021swin} to introduce connections across windows at each stage of the network) is done by displacing the windows by $M$ patches along the azimuth.


\subsection{Angular relative positional encoding}
\label{sec:positional-encoding}

The Swin transformer relative positional encoding~\cite{liu2021swin} includes a position bias $B \in \mathbb{R}^{M^2 \times M^2}$, where $M^2$ is the number of patches in a window, added to each head when computing similarity:
\begin{equation}
\text{Att}(Q,K,V) = \text{Softmax}(QK^T / \sqrt{d} + B)V \,,
\label{eq:attnswin}
\end{equation}
where $Q,K,V$ $\in$ $\mathbb{R}^{M^2 \times d}$ are the queries, keys, and values matrices respectively; and $d$ is the query/key dimension. 

Since \thename follows a radial partitioning, we employ an angular relative positional encoding to capture the relative position between tokens with respect to incident $\theta$ and azimuth $\varphi$ angles.
We divide the position bias $B$ in \cref{eq:attnswin} into two parts: $B_\theta$ and $B_\varphi$, the incident-angular and azimuthal relative position bias respectively. Given the $i$-th token angular coordinates 
\begin{equation}
\theta_i = \frac{\theta_\mathrm{max} (i-0.5)}{N_r} \; \text{and} \;
\varphi_i = \frac{2\pi (i-0.5)}{N_\varphi} \,,
\end{equation}
where $\theta_\mathrm{max}$ is the half field of view. The relative angular positions $(\Delta \theta, \Delta \varphi)$ between tokens $i,j$ are given by
\begin{align}
\Delta \theta &= \theta_i - \theta_j \,, \text{where} \, i,j \in [1, ..., N_\theta] \,, \text{and} \nonumber \\
\Delta \varphi &= \varphi_i- \varphi_j \,, \text{where} \, i,j \in [1, ..., N_\varphi] \,.
\end{align}
%
The two tensors $B_\varphi$ and $B_\theta$ are defined as
%
\begin{align}
  B_\theta &= a_{\Delta \theta} \sin(\Delta \theta) + b_{\Delta \theta} \cos(\Delta \theta) \,, \\
B_\varphi &= a_{\Delta \varphi} \sin(\Delta \varphi) + b_{\Delta \varphi} \cos(\Delta \varphi) \,.
  \label{eq:positional-encoding}
\end{align}
Here, $a_*$ and $b_*$ are trainable coefficients. Since the relative positions in a window along angular and azimuth axes ranges from $[-M_\theta + 1, M_\theta - 1]$ and $[-M_\varphi + 1, M_\varphi - 1]$ respectively, where $M_\theta$ and $M_\varphi$ are number of patches in a window (see \cref{sec:window-attn}, here $M_\theta=M_r$). We parameterize two bias matrices $\hat{B}_\varphi \in \mathbb{R}^{(2M_\varphi - 1) \times 2}$ and $\hat{B}_\theta \in \mathrm{R}^{(2M_\theta - 1) \times 2}$. Hence $a_*$ and $b_*$ are taken from $\hat{B}_\theta$ and $\hat{B}_\varphi$.


Finally, the two tensors $B_\theta,B_\varphi \in \mathbb{R}^{M_r^2 \times M_\varphi^2}$ are built on all pairs of tokens. The final attention equation is thus
\begin{equation}
    \text{Att}(Q,K,V) = \text{Softmax}(QK^T / \sqrt{d} + B_\theta + B_\varphi)V \,.
    \label{eq:attention}
\end{equation}


\subsection{Polar patch merging}
\label{sec:patch-merging}

Similar to window-based self attention (\cref{sec:window-attn}), the polar nature of our architecture enables many possibilities when merging patches. For example, we could merge $2 \times 2$ neighboring patches (\thename-RA) or $1 \times 4$ merge along azimuth (\thename-A) as shown in \cref{tab:detailed architecture}. We found the azimuth merging strategy to outperform the others in our experiments (see \cref{sec:ablation}).

\begin{table}[]
\footnotesize
\centering
\caption{\Ak{Model architecture specification and variants. In both variants, we adapt the small architecture of Swin Transformer \cite{liu2021swin}, which has four stages of patch merging. \textbf{Input Resolution} : Resolution of feature map at every stage. \textbf{Window Size} : Window size ($M_r, M_\varphi$), $M_r$ : Number of patches along radial axis in a window, $M_\varphi$ : Number of patches along azimuth axis in a window. \textbf{DarSwin-RA}: Architecture variant with four patches along both azimuth and radial axis and merging $2 \times 2$ neighborhood patches along the radius and azimuth. \textbf{DarSwin-A}: architecture variant with 16 patches along the azimuth axis and one patch along the radial axis in a window and merging $1 \times 4$ neighborhood patches along the radius and azimuth.}}
\begin{tabular}{cccc}
\hline
\multicolumn{2}{c}{DarSwin-RA}                                                                                                                                      & \multicolumn{2}{c}{DarSwin-A}                                                                                                                                       \\ \hline
\begin{tabular}[c]{@{}c@{}}Input Resolution\end{tabular} & \begin{tabular}[c]{@{}c@{}}Window Size\\ $(M_r, M_\varphi)$\end{tabular} & \begin{tabular}[c]{@{}c@{}}Input Resolution\end{tabular} & \begin{tabular}[c]{@{}c@{}}Window Size\\ $(M_r, M_\varphi)$\end{tabular} \\ \hline
\begin{tabular}[c]{@{}c@{}}Stage-1\\ (16, 64)\end{tabular}                & (4, 4)                                                                                  & \begin{tabular}[c]{@{}c@{}}Stage-1\\ (16, 64)\end{tabular}                & (1, 16)                                                                                 \\ \hline
\begin{tabular}[c]{@{}c@{}}Stage-2\\ (8, 32)\end{tabular}                 & (4, 4)                                                                                  & \begin{tabular}[c]{@{}c@{}}Stage-2\\ (16, 16)\end{tabular}                & (1, 16)                                                                                 \\ \hline
\begin{tabular}[c]{@{}c@{}}Stage-3\\ (4, 16)\end{tabular}                 & (4, 4)                                                                                  & \begin{tabular}[c]{@{}c@{}}Stage-3\\ (16, 4)\end{tabular}                 & (4, 4)                                                                                  \\ \hline
\begin{tabular}[c]{@{}c@{}}Stage-4\\ (2, 8)\end{tabular}                  & (2, 8)                                                                                  & \begin{tabular}[c]{@{}c@{}}Stage-4\\ (16, 1)\end{tabular}                 & (16, 1)                                                                                 \\ \hline
\end{tabular}
\label{tab:detailed architecture}
\end{table}

\section{Classification experiments}
\label{sec:experiments}

To evaluate the efficacy of our proposed \thename encoder, we perform a series of experiments on image classification. Since there exists no classification dataset for wide angle images, we instead create synthetically distorted images using 200 randomly chosen classes from the ImageNet1k dataset~\cite{krizhevsky2012classif}. 


\subsection{Dataset}
\label{sec:dataset}

\begin{figure}
\footnotesize
\setlength{\tabcolsep}{1pt}
\begin{tabular}{cccc}
Very low & Low & Medium & High \\
\includegraphics[width = 0.24\linewidth]{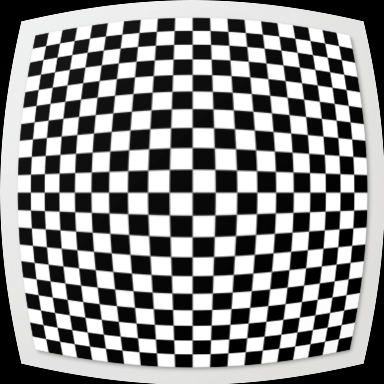} &
\includegraphics[width = 0.24\linewidth]{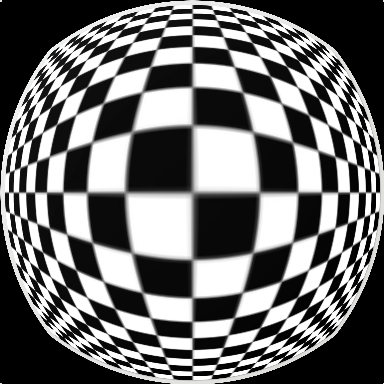} &
\includegraphics[width = 0.24\linewidth]{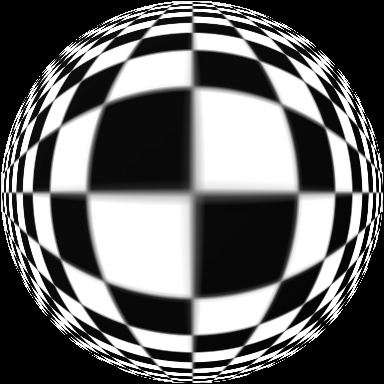} &
\includegraphics[width = 0.24\linewidth]{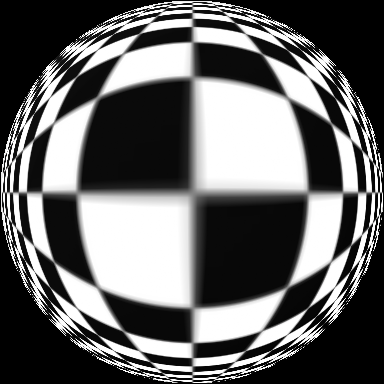} \\
\includegraphics[width = 0.24\linewidth]{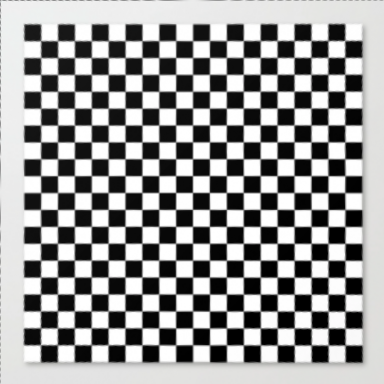} &
\includegraphics[width = 0.24\linewidth]{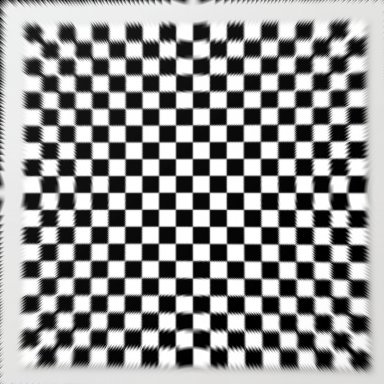} &
\includegraphics[width = 0.24\linewidth]{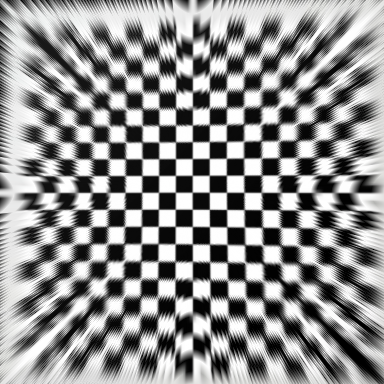} &
\includegraphics[width = 0.24\linewidth]{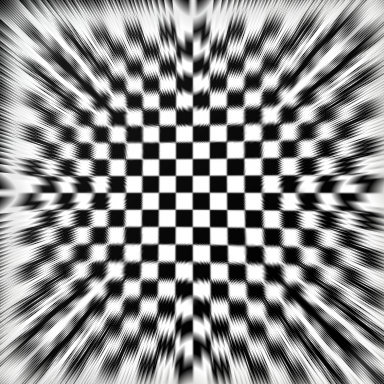}
\end{tabular}
\caption{Visualization of a checkerboard pattern distorted according to each distortion level at its original resolution of (224, 224) and downsampled to (64, 64) in our four training sets. From left to right: very low, low, medium, and high. The second row represents the respective undistorted images.}
\label{fig:distortion-levels}
\end{figure}

To evaluate our approach on a wide range of conditions, we employ the unified spherical projection model (c.f. \cref{sec:background}) to synthetically distort the (perspective) ImageNet1k images. For this, we warp the images at the original pixel resolution ($224 \times 224$) then downsample to ($64 \times 64$) for all experiments.
\paragraph{Training sets} We generate four different training sets withdifferent levels of distortion, defined by the distortion parameter $\xi$: ``very low'' ($\xi \in [0.0, 0.05]$), ``low'' ($\xi \in [0.2, 0.35]$), ``medium'' ($\xi \in [0.5, 0.7]$), and ``high'' ($\xi \in [0.85, 1.0]$). \Cref{fig:distortion-levels} shows examples of a checkerboard image distorted at each level. Training images are distorted on-the-fly during training with the distortion level sampled uniformly from the interval mentioned above. Each training set contains 260,000 images and 10,000 validation images, over 200 classes. 
\paragraph{Test set} Test sets of 30,000 images over 200 classes are generated using the same procedure. Here, $\xi$ is determined once for each image and kept fixed. And we test for all $\xi$ values between $[0, 1]$ to evaluate generalization to different lens distortions.




\subsection{Baselines and training details}
\label{sec:baselines}

We compare our approach with the following baselines: Swin-S~\cite{liu2021swin}, Deformable Attention Transformer (DAT-S)~\cite{xia2022vision}, and Swin-S on input undistorted images dubbed ``Swin(undis)''. As with our DarSwin, this last baseline has knowledge of the distortion parameters whereas the first two do not. Note that we do not include comparisons to methods which estimate distortion \cite{yin2018fisheyerectnet, xue2019learning, 9980359, zhang2015line}. Indeed, the spherical projection model (\cref{sec:background}) is bijective: the undistortion function has a closed form and is exact. Therefore, the ``Swin (undis)'' method serves as an upper bound to all self-calibration methods because it is, in essence, being given the ``ground truth'' undistortion.

All three baselines employ 32 divisions along the image width and height. For DarSwin, we use $N_r = 16$ and $N_\varphi = 64$ divisions along the radius and azimuth respectively, which yields the same total number of 1024 patches for all methods. All three baselines are trained with a window size (4, 4) on our synthetically distorted training sets.

All methods use the AdamW optimizer on a batch size of 128 using a cosine decay learning rate scheduler and 20 epochs of linear warm-up. We use an initial learning rate of 0.001 and a weight decay of 0.05. We include all of the augmentation and regularization strategies of \cite{liu2021swin}, except for random crop and geometric transformation (like shearing and translation). Our model requires 0.03M additional parameters over the Swin baseline which contains 48M, representing a 0.061\% increase.



\begin{figure*}[!h]
\centering
\footnotesize
\setlength{\tabcolsep}{1pt}
\begin{tabular}{cccc}
\includegraphics[width=0.38\linewidth,trim=0 0 1cm 0,clip]{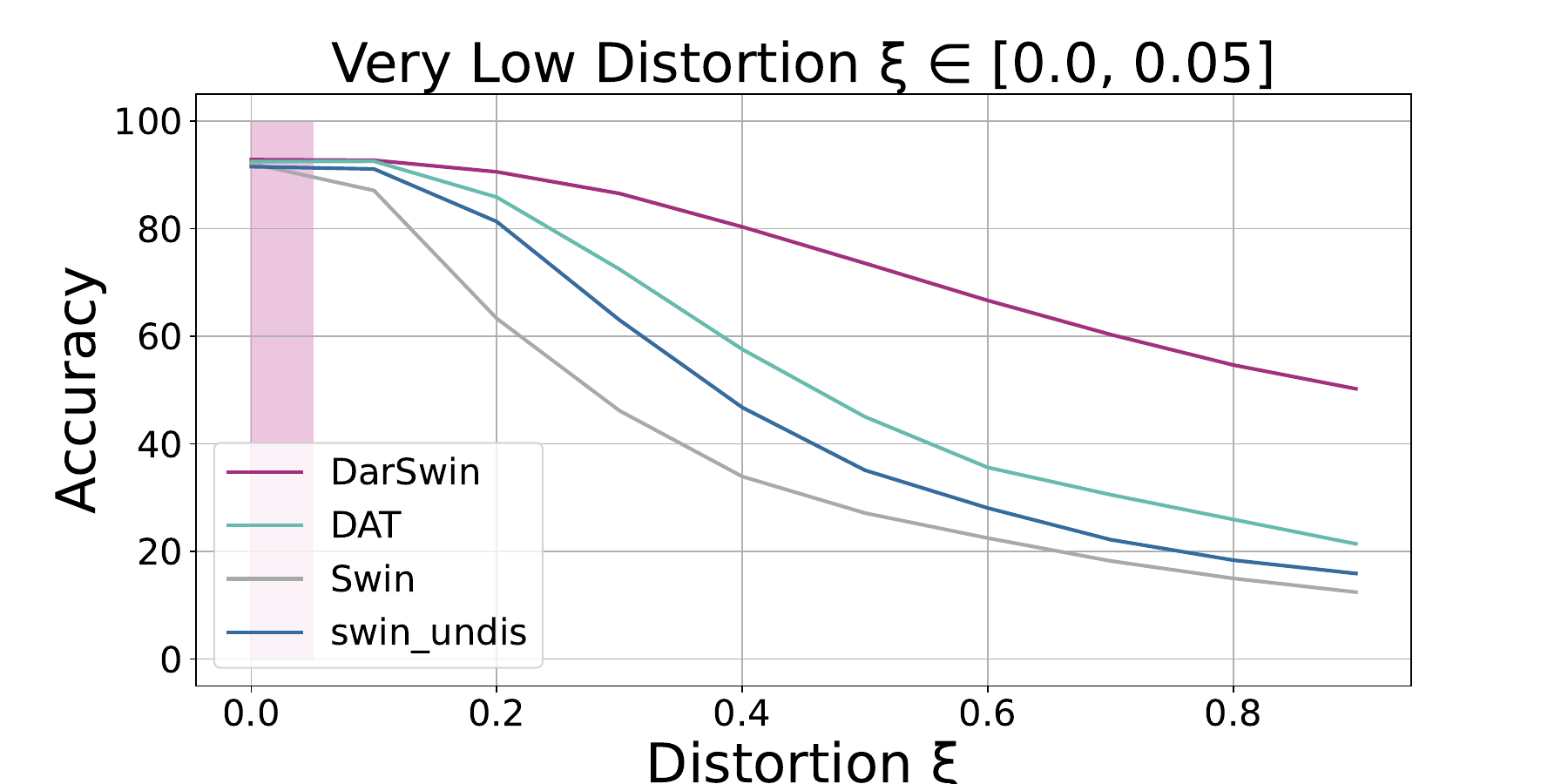} & 
\hspace{10pt}
\includegraphics[width=0.38\linewidth,trim=0 0 1cm 0,clip]{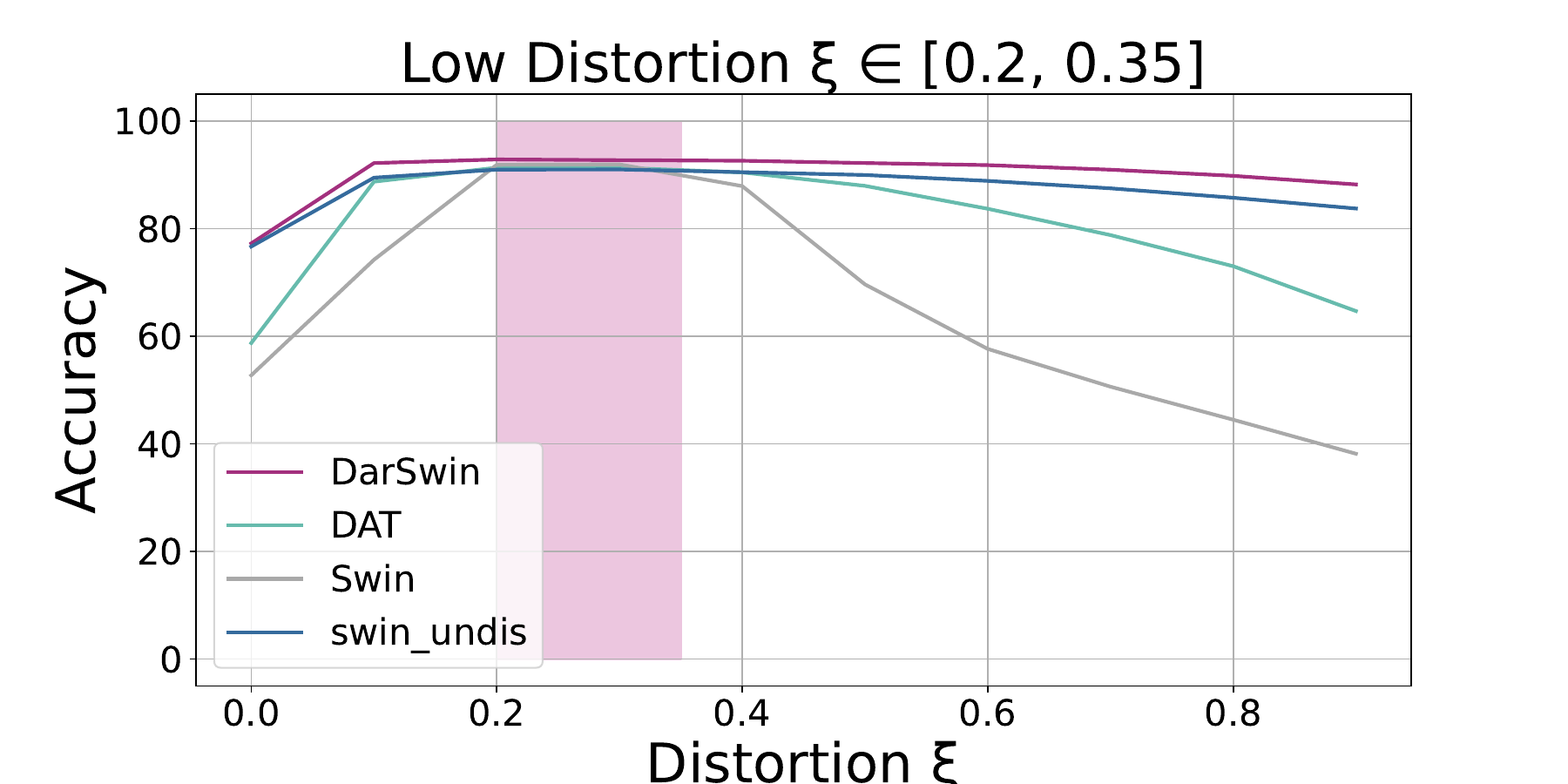} \\
(a) Very low & (b) Low \\
\includegraphics[width=0.38\linewidth,trim=0 0 1cm 0,clip]{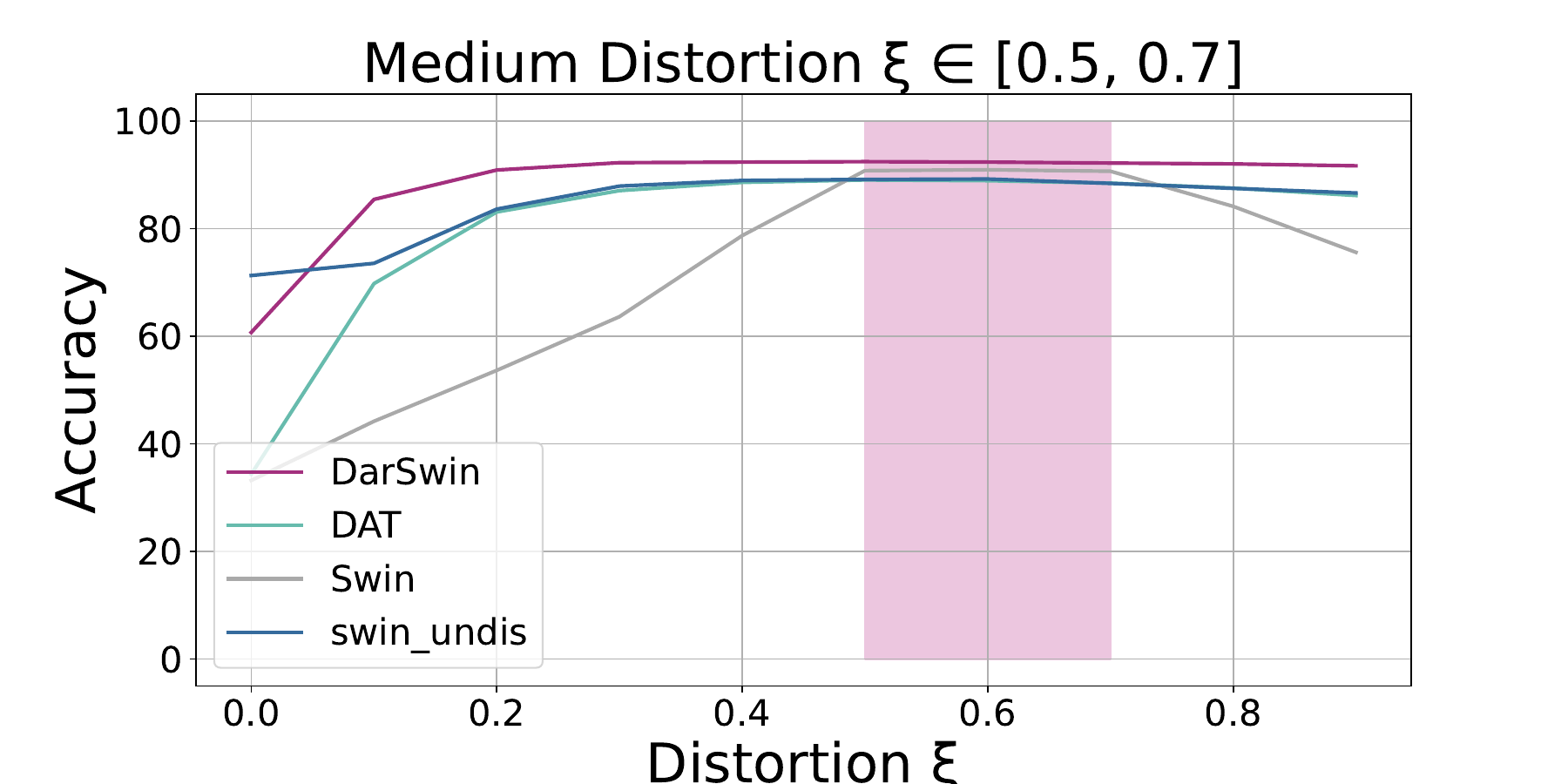} & 
\hspace{10pt}

\includegraphics[width=0.38\linewidth,trim=0 0 1cm 0,clip]{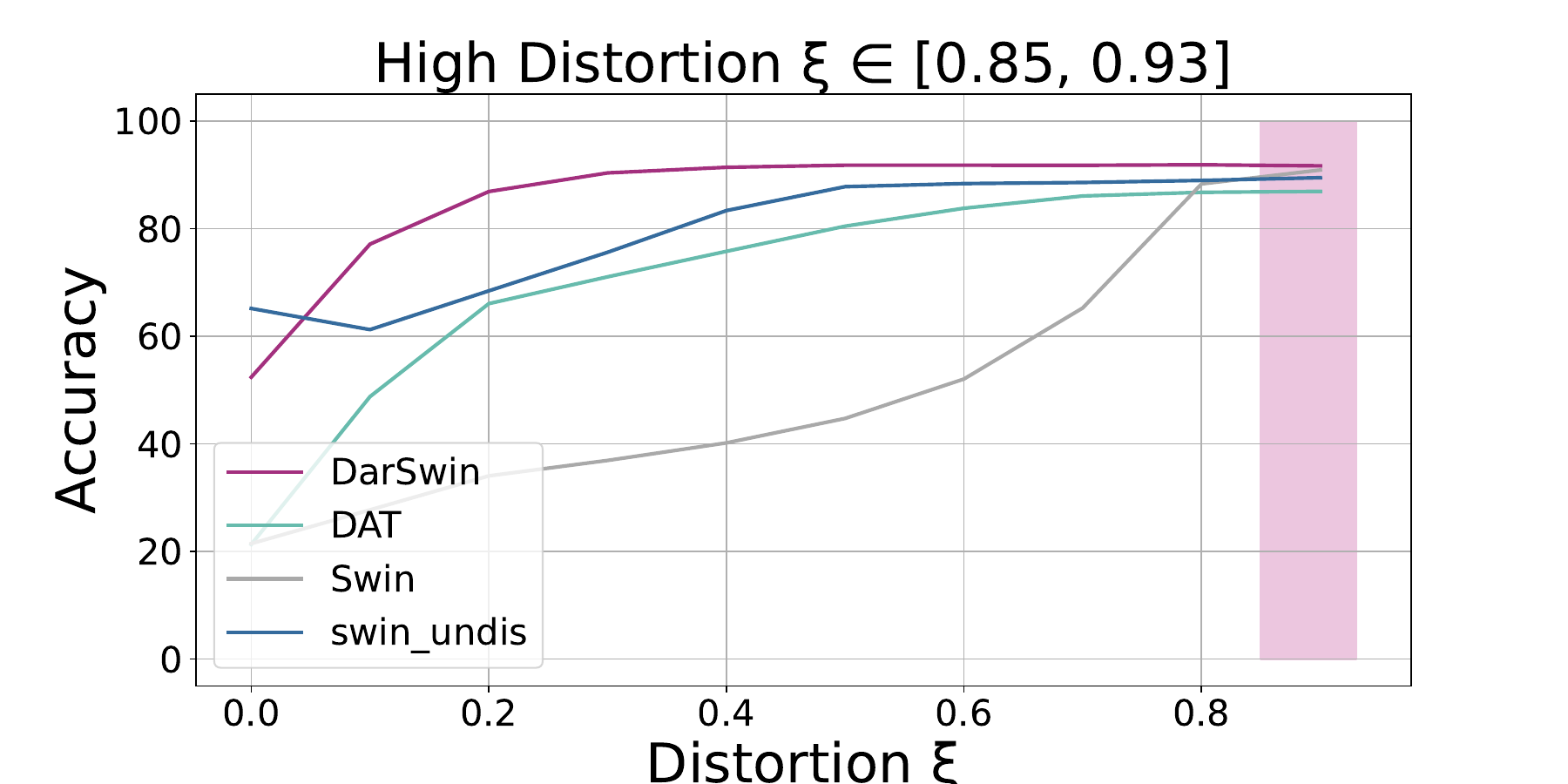} \\
(c) Medium & (d) High
\end{tabular}
\caption[]{Top-1 classification accuracy (mean) as a function of test distortion for our method (\thename-A) and previous state of the art: DAT~\cite{xia2022vision}, Swin~\cite{liu2021swin}, and Swin (undis) (see text). All methods are trained on a restricted set of lens distortion curves (indicated by the pink shaded regions): (a) Very low, (b) low, (c) medium and (d) high distortion. We observe zero-shot adaptation to lens distortion of each approach by testing across all $\xi \in [0, 1]$.}
\label{fig:results-distortion}
\end{figure*}



\subsection{Zero-shot lens distortion generalization} 
\label{sec:generalization-results}

We are interested in evaluating whether our distortion-aware method can better generalize to other unseen lenses at test time. Hence, we train each approach on a level of distortion (c.f. \cref{sec:dataset}) and evaluate them on all distortion values $\xi \in [0, 1]$. For this, entire test set is distorted using a single $\xi$ value, and we repeat this process for every $\xi \in [0, 1]$ to simulate different lens distortion. 

Results are reported in \cref{fig:results-distortion}, where the distribution of training distortions is drawn in pink. We observe that \thename performs on par with the baselines when test distortions overlap with training, but shows much greater generalization capabilities outside of the training domain. Furthermore, \cref{tab:distortion_results_table} shows the top-1 accuracy for the test set distorted with $\xi = 0.4$, which is in none of the training intervals. Again, we note that \thename yields better generalization accuracy (without fine-tuning) than all baselines, even Swin (undis) which also has access to the ground truth distortion function (see \cref{sec:baselines}). 

\begin{table}[!h]
\centering
\footnotesize
\caption{Comparisons of top-1 accuracy of the models trained on four levels of distortion (pink regions in \cref{fig:results-distortion}) and tested on distortion level $\xi = 0.4$. Each row is color-coded as \colorbox{best}{best} and \colorbox{second}{second best}.}
\begin{tabular}{@{}ccccc@{}}
\toprule
Methods      & Very Low & Low  & Medium & High \\ \midrule
DarSwin      & \cellcolor{best}{80.33}   & \cellcolor{best}{92.61}  & \cellcolor{best}{92.35}   & \cellcolor{best}{91.39} \\
Swin     & 33.94     & 87.90 & 78.7   & 40.1 \\
Swin (undis) & 47.48     & \cellcolor{second}{91.52} & \cellcolor{second}{91.07}   & \cellcolor{second}{87.34} \\
DAT          &  \cellcolor{second}{57.5}  & 90.4 & 88.5   & 75.7 \\ \bottomrule
\end{tabular}
\label{tab:distortion_results_table}
\end{table}



\subsection{Ablations}
\label{sec:ablation}

We ablate some important design elements (\cref{tab:ablation}) and training strategies (\cref{tab:ablation_train}). All ablations are performed using \thename-A trained on the ``low'' distortion ($\xi \in [0.2, 0.35]$) and tested on all the distortion levels. 


\paragraph{Positional encoding (PE)} We compare our angular relative position encoding (\cref{sec:positional-encoding}) with two versions of the Fourier-based polar positional encoding from \cite{ning2020polar} in the first part of \cref{tab:ablation}. First, polar PE encodes patches using their relative distortion-aware radial lengths $r$ and azimuth values $\varphi$. Second, angular PE encodes patches using the relative value of incident angle $\theta$ and azimuth $\varphi$. While angular PE yields better performance, we observe that our angular relative PE improves the performance even further. 


\paragraph{Number of sampling points along radius $S_r$} 
We observe in \cref{tab:ablation} that \thename with $S_r=10$ sampling points along the radius outperforms the models with $S_r = 5$ and $S_r = 2$ sampling points along the radius by just over (1-6)\%.

\paragraph{Window formations and merging} Ablations of different window formation and merging strategies : along azimuth (DarSwin-A) or along radius+azimuth (DarSwin-RA) respectively are reported in \cref{tab:ablation}. We observe that merging along the azimuth outperforms the merging along randius+azimuth by approximately (1-4)\% on each distortion level.

\begin{table}
\centering
\footnotesize
\caption{Ablation study on different design elements of \thename, including different positional encoding, number of sampling points along the radius $n_r$, and window formation and merging strategy: \thename-A (azimuth) or \thename-RA (radius+azimuth ).}
\begin{tabular}{lcccc}
\toprule
              & Very low & Low & Medium & High \\
\midrule
Angular relative PE  &        83.43\%        &        92.8\%        &       91.5\%            &       88.27\%  \\
Polar PE   \cite{ning2020polar}    &       79.6\%    &      92.0\%     & 90.1\%     &   85.4\%              \\
Angular PE \cite{ning2020polar}&    81.47\%           &       92.617\%         &         91.11\%          &      87.4\%           \\
\midrule
$S_r$ = 10  &        83.43\%        &        92.8\%        &       91.5\%            &       88.27\%  \\
$S_r$ = 5    &          81.81\%     &      91.94\%          &        90.00\%         &    85.19\%              \\
$S_r$ = 2    &      78.9\%         &      90.4\%        &        87.9\%           &         82.4\%        \\
\midrule
DarSwin-A &        83.43\%        &        92.8\%        &       91.5\%            &       88.27\%  \\
DarSwin-RA  &  \Ak{79.7\%}  & \Ak{92.7\%}    & \Ak{90.3\%}   & \Ak{84.7\%}                             \\   
\bottomrule
\end{tabular}
\label{tab:ablation}
\end{table}

\begin{table}
\centering
\footnotesize
\caption{\Ak{Ablation study on different types of sampling techniques and augmentation strategies on DarSwin-A. For jittering we compare ``no jittering'' (all samples are given to the MLP for linear embedding as is) with ``with jittering''( jittering is applied on the sample points in a patch). For sampling, we compare ``Uniform sampling'': points are sampled uniformly inside a patch; to ``Distortion aware (DA) sampling'': lens information is taken into account to sample points inside a patch. "Distortion aware (DA) partition" DarSwin uses lens information to partition the patch.}}
\resizebox{\columnwidth}{!}{%
\begin{tabular}{lcccc} 
\toprule
    & Very low & Low & Medium & High  \\ 
\midrule
With jittering       & 83.43\%        &        92.8\%        &       91.5\%            &       88.27\%  \\
No jittering     & 81.5\%    & 92.6\%    & 91.16\%   & 87.4 \% \\ 
\midrule
DA sampling &  83.43\%        &        92.8\%        &       91.5\%            &       88.27\%  \\
Uniform sampling          &  82.4\%   & 92.5\%    & 91.2\%   & 86.2\% \\ 
\midrule
\Ak{DA partition}          &  83.43\%        &        92.8\%        &       91.5\%            &       88.27\%  \\
\Ak {w/o DA partition} & \Ak{52.2\%}  & \Ak{91.3\%}    & \Ak{86.2\%}   & \Ak{75.5\%}                             \\
\bottomrule
\end{tabular}%
}
\label{tab:ablation_train}
\end{table}


\paragraph{Jittering.} As discussed in \cref{sec:linear-embedding}, when patches are sampled for the linear embedding layer, an augmentation strategy is used to jitter the sample points in the patch. According to \cref{tab:ablation_train}, this jittering augmentation improves the performance by (1--2)\%.

\paragraph{Distortion aware (DA) sampling.} We observe in \cref{tab:ablation_train} that distortion aware sampling improves performance by almost (2--6)\% compared to uniform sampling in a patch.

\paragraph{Distortion aware (DA) patch partition.} We observe in \cref{tab:ablation_train}  that without lens information, DarSwin cannot generalize to unseen distortion.

\subsection{Generalization over projection models}

In \cref{tab:poly}, we aim to check for
robustness for domain shift at inference due to the use of
a different distortion model (c.f. \cref{sec:background}). In particular, we use the test set of 30,000 images and distort them using the distortion parameters of a 4-degree polynomial projection model. Each image in the test set is assigned four different distortion parameters randomly sampled from a uniform distribution. The undistortion is not exact since the polynomial projection function is not invertible. Hence Swin (undis) performs only better than DarSwin on low distortions, but the performance degrades on medium to high distortion levels. Our model performs better generalization on all distortion levels. DAT fails to generalize on low distortion levels, and Swin (undis) fails to generalize on high distortions.

\begin{table}[]
\centering
\footnotesize
\caption{Comparison of our method with baselines on generalization across projection model. We record top-1 accuracy on the polynomial projection test set for all methods trained on four distortion levels of the spherical projection model. Each row is color-coded as \colorbox{best}{best} and \colorbox{second}{second best}.}
\begin{tabular}{ccccc}
\toprule
Training Levels & Very Low & Low    & Medium & High   \\ \hline
Swin(undis)     & \cellcolor{best}{85.9\%} &\cellcolor{best}{87.1\%} & 71.3\% & 60.3\% \\ 
DAT             & 65.4\%   & \cellcolor{second}{85.9\%} & \cellcolor{best}{78.9\%} & \cellcolor{second}{63.8\%} \\
Swin            & 33.5\%   & 70.2\% & 42.3\% & 25.6\% \\
Ours            & \cellcolor{second}{82.7\%}   & 84.8\% & \cellcolor{second}{78.4\%} & \cellcolor{best}{65.5\%}   \\ \hline
\end{tabular}
\label{tab:poly}
\end{table}

\section{Discussion}
\label{sec:Discussion}

This paper presents \thename, a new distortion aware vision transformer which adapts its structure to the lens distortion profile of a (calibrated) lens. \thename achieves state-of-the-art performance on zero-shot adaptation (without pretraining) on different lenses on classification using synthetically distorted images from the ImageNet1k dataset. 
\paragraph{Limitations and future research directions} While our method demostrates state-of-the-art performance, it suffers from some limitations. First, the distortion aware sampling strategy is shown to be effective for zero-shot adaptation, the sparsity of sampling points and necessity to interpolate pixel values may affect the performance of the model. While this issue is partially alleviated using our proposed jittering augmentation technique, other strategies may also be possible. Second, our model assumes knowledge of the lens distortion profile, hence it is appropriate only for the calibrated case. We hope to extend our work to uncalibrated lenses, for example by taking inspiration from \cite{Hold-Geoffroy_2018_CVPR, yin2018fisheyerectnet, article1}. Finally, while our experiments demonstrate promising performance on classification experiments, we wish to expand to per-pixel tasks, such as semantic classification and depth estimation. Here, making pixel decoders distortion aware is an exciting direction for future work. 


\paragraph{Acknowledgments} This research was supported by NSERC grant ALLRP-567654, Thales, an NSERC USRA to J. Lagüe, and the Digital Research Alliance Canada. We thank Yohan Poirier-Ginter, Frédéric Fortier-Chouinard, Adam Tupper and Justine Giroux for proofreading.
{
\bibliographystyle{ieee_fullname}
\bibliography{bibliography}

\begin{thebibliography}{10}\itemsep=-1pt

\bibitem{ahmad2022fisheyehdk}
Ola Ahmad and Freddy Lecue.
\newblock {FisheyeHDK}: Hyperbolic deformable kernel learning for ultra-wide
  field-of-view image recognition.
\newblock In {\em AAAI}, 2022.

\bibitem{barreto2006unifying}
Jo{\~a}o~P. Barreto.
\newblock A unifying geometric representation for central projection systems.
\newblock {\em Comput. Vis. Img. Underst.}, 2006.

\bibitem{beck1925apparatus}
Conrad Beck.
\newblock Apparatus to photograph the whole sky.
\newblock {\em J. Scientific Inst.}, 2(4):135--139, 1925.

\bibitem{GDL}
Michael~M. Bronstein, Joan Bruna, Taco Cohen, and Petar Velickovic.
\newblock Geometric deep learning: Grids, groups, graphs, geodesics, and
  gauges.
\newblock {\em CoRR}, abs/2104.13478, 2021.

\bibitem{brousseau2019calibration}
Pierre-Andre Brousseau and Sebastien Roy.
\newblock Calibration of axial fisheye cameras through generic virtual central
  models.
\newblock In {\em Int. Conf. Comput. Vis.}, 2019.

\bibitem{cohen2018spherical}
Taco~S. Cohen, Mario Geiger, Jonas K{\"{o}}hler, and Max Welling.
\newblock Spherical cnns.
\newblock In {\em Int. Conf. Learn. Represent.}, 2018.

\bibitem{cohen2019gauge}
Taco~S. Cohen, Maurice Weiler, Berkay Kicanaoglu, and Max Welling.
\newblock Gauge equivariant convolutional networks and the icosahedral {CNN}.
\newblock In {\em Int. Conf. on Mach. Learning}, 2019.

\bibitem{dai2017deformable}
Jifeng Dai, Haozhi Qi, Yuwen Xiong, Yi Li, Guodong Zhang, Han Hu, and Yichen
  Wei.
\newblock Deformable convolutional networks.
\newblock In {\em Int. Conf. Comput. Vis.}, 2017.

\bibitem{deng2019restricted}
Liuyuan Deng, Ming Yang, Hao Li, Tianyi Li, hu Bing, and Chunxiang Wang.
\newblock Restricted deformable convolution-based road scene semantic
  segmentation using surround view cameras.
\newblock {\em IEEE Trans. Int. Trans. Syst.}, 08 2019.

\bibitem{article1}
Frédéric Devernay and Olivier Faugeras.
\newblock Straight lines have to be straight automatic calibration and removal
  of distortion from scenes of structured environments.
\newblock {\em Mach. Vis. Appl.}, 13, 08 2001.

\bibitem{dosovitskiy2020vit}
Alexey Dosovitskiy, Lucas Beyer, Alexander Kolesnikov, Dirk Weissenborn,
  Xiaohua Zhai, Thomas Unterthiner, Mostafa Dehghani, Matthias Minderer, Georg
  Heigold, Sylvain Gelly, Jakob Uszkoreit, and Neil Houlsby.
\newblock An image is worth 16x16 words: Transformers for image recognition at
  scale.
\newblock In {\em Int. Conf. Learn. Represent.}, 2020.

\bibitem{fernandez2018layouts}
Clara Fernandez-Labrador, Alejandro Perez-Yus, Gonzalo Lopez-Nicolas, and
  Jose~J Guerrero.
\newblock Layouts from panoramic images with geometry and deep learning.
\newblock {\em IEEE Rob. Autom. Letters}, 3(4):3153--3160, 2018.

\bibitem{fleck1995perspective}
Margaret~M. Fleck.
\newblock Perspective projection: The wrong imaging model.
\newblock {\em IEEE Trans. Reliability}, 1995.

\bibitem{hill1924lens}
Robin Hill.
\newblock A lens for whole sky photographs.
\newblock {\em Quart. J. Royal Meteo. Soc.}, 50(211):227--235, 1924.

\bibitem{Hold-Geoffroy_2018_CVPR}
Yannick Hold-Geoffroy, Kalyan Sunkavalli, Jonathan Eisenmann, Matthew Fisher,
  Emiliano Gambaretto, Sunil Hadap, and Jean-François Lalonde.
\newblock A perceptual measure for deep single image camera calibration.
\newblock In {\em Proceedings of the IEEE Conference on Computer Vision and
  Pattern Recognition (CVPR)}, June 2018.

\bibitem{huang2018multimodal}
Xun Huang, Ming-Yu Liu, Serge Belongie, and Jan Kautz.
\newblock Multimodal unsupervised image-to-image translation.
\newblock In {\em Eur. Conf. Comput. Vis.}, 2018.

\bibitem{hughes2010accuracy}
Ciar{\'a}n Hughes, Patrick Denny, Edward Jones, and Martin Glavin.
\newblock Accuracy of fish-eye lens models.
\newblock {\em Applied optics}, 49(17):3338--3347, 2010.

\bibitem{jang2022dada}
Sujin Jang, Joohan Na, and Dokwan Oh.
\newblock Dada: Distortion-aware domain adaptation for unsupervised semantic
  segmentation.
\newblock In {\em Adv. Neural Inform. Process. Syst.}, 2022.

\bibitem{kannala2006generic}
Juho Kannala and Sami~S. Brandt.
\newblock A generic camera model and calibration method for conventional,
  wide-angle, and fish-eye lenses.
\newblock {\em IEEE Trans. Pattern Anal. Mach. Intell.}, 22(8):1335--1340,
  2006.

\bibitem{khosla2012undoing}
Aditya Khosla, Tinghui Zhou, Tomasz Malisiewicz, Alexei~A Efros, and Antonio
  Torralba.
\newblock Undoing the damage of dataset bias.
\newblock In {\em Eur. Conf. Comput. Vis.}, 2012.

\bibitem{9980359}
Byunghyun Kim, Dohyun Lee, Kyeongyuk Min, Jongwha Chong, and Inwhee Joe.
\newblock Global convolutional neural networks with self-attention for fisheye
  image rectification.
\newblock {\em IEEE Access}, 10:129580--129587, 2022.

\bibitem{kim2015fisheye}
Hyungtae Kim, Eunjung Chae, Gwanghyun Jo, and Joonki Paik.
\newblock Fisheye lens-based surveillance camera for wide field-of-view
  monitoring.
\newblock In {\em IEEE Int. Conf. Cons. Elec.}, 2015.

\bibitem{krizhevsky2012classif}
Alex Krizhevsky, Ilya Sutskever, and Geoffrey~E Hinton.
\newblock Imagenet classification with deep convolutional neural networks.
\newblock In {\em Adv. Neural Inform. Process. Syst.}, 2012.

\bibitem{kumar2022surround}
Varun~Ravi Kumar.
\newblock Surround-view cameras based holistic visual perception for automated
  driving.
\newblock {\em arXiv preprint arXiv:2206.05542}, 2022.

\bibitem{kumar2021omnidet}
Varun~Ravi Kumar, Senthil Yogamani, Hazem Rashed, Ganesh Sitsu, Christian Witt,
  Isabelle Leang, Stefan Milz, and Patrick M{\"a}der.
\newblock Omnidet: Surround view cameras based multi-task visual perception
  network for autonomous driving.
\newblock {\em IEEE Robot. Autom. Letters}, 6(2):2830--2837, 2021.

\bibitem{liao2022kitti}
Yiyi Liao, Jun Xie, and Andreas Geiger.
\newblock Kitti-360: A novel dataset and benchmarks for urban scene
  understanding in 2d and 3d.
\newblock {\em IEEE Trans. Pattern Anal. Mach. Intell.}, 2022.

\bibitem{googlenet}
Shuying Liu and Weihong Deng.
\newblock Very deep convolutional neural network based image classification
  using small training sample size.
\newblock In {\em 2015 3rd IAPR Asian Conference on Pattern Recognition
  (ACPR)}, 2015.

\bibitem{liu2021swin}
Ze Liu, Yutong Lin, Yue Cao, Han Hu, Yixuan Wei, Zheng Zhang, Stephen Lin, and
  Baining Guo.
\newblock Swin transformer: Hierarchical vision transformer using shifted
  windows.
\newblock In {\em Int. Conf. Comput. Vis.}, 2021.

\bibitem{mei2007single}
Christopher Mei and Patrick Rives.
\newblock Single view point omnidirectional camera calibration from planar
  grids.
\newblock In {\em Int. Conf. Robot. Aut.}, 2007.

\bibitem{melo2013unsupervised}
R. Melo, M. Antunes, J.~P. Barreto, G. Falcão, and N. Gonçalves.
\newblock Unsupervised intrinsic calibration from a single frame using a
  ``plumb-line`` approach.
\newblock In {\em Int. Conf. Comput. Vis.}, 2013.

\bibitem{miyamoto1964fisheye}
Kenro Miyamoto.
\newblock Fish eye lens.
\newblock {\em J. Opt. Soc. Am.}, 54(8):1060--1061, Aug 1964.

\bibitem{ning2020polar}
Ke Ning, Lingxi Xie, Fei Wu, and Qi Tian.
\newblock Polar relative positional encoding for video-language segmentation.
\newblock In {\em Int. Joint Conf. Art. Intel.}, 2020.

\bibitem{plaut2021fisheye}
Elad Plaut, Erez Ben~Yaacov, and Bat El~Shlomo.
\newblock 3d object detection from a single fisheye image without a single
  fisheye training image.
\newblock In {\em IEEE Conf. Comput. Vis. Pattern Recog. Worksh.}, 2021.

\bibitem{playout2021adaptable}
Cl{\'{e}}ment Playout, Ola Ahmad, Freddy L{\'{e}}cu{\'{e}}, and Farida Cheriet.
\newblock Adaptable deformable convolutions for semantic segmentation of
  fisheye images in autonomous driving systems.
\newblock {\em CoRR}, abs/2102.10191, 2021.

\bibitem{ramalingan2017unifying}
S. Ramalingam and Peter Sturm.
\newblock A unifying model for camera calibration.
\newblock {\em IEEE Trans. Pattern Anal. Mach. Intell.}, 39(7):1309--1319,
  2017.

\bibitem{Rashed_2021_WACV}
Hazem Rashed, Eslam Mohamed, Ganesh Sistu, Varun~Ravi Kumar, Ciaran Eising,
  Ahmad El-Sallab, and Senthil Yogamani.
\newblock Generalized object detection on fisheye cameras for autonomous
  driving: Dataset, representations and baseline.
\newblock In {\em Wint. Conf. Appl. Comput. Vis.}, 2021.

\bibitem{inproceedings}
Hazem Rashed, Eslam Mohamed, Ganesh Sistu, Varun Ravi~Kumar, Ciaran Eising,
  Ahmad Sallab, and Senthil Yogamani.
\newblock Fisheyeyolo: Object detection on fisheye cameras for autonomous
  driving.
\newblock In {\em Adv. Neural Inform. Process. Syst.}, 12 2020.

\bibitem{schmalstieg2017ar}
Dieter Schmalstieg and Tobias Höllerer.
\newblock Augmented reality: Principles and practice.
\newblock In {\em IEEE Virt. Reality}, 2017.

\bibitem{su2019kernel}
Yu-Chuan Su and Kristen Grauman.
\newblock Kernel transformer networks for compact spherical convolution.
\newblock In {\em IEEE Conf. Comput. Vis. Pattern Recog.}, 2019.

\bibitem{VGG}
Christian Szegedy, Wei Liu, Yangqing Jia, Pierre Sermanet, Scott~E. Reed,
  Dragomir Anguelov, Dumitru Erhan, Vincent Vanhoucke, and Andrew Rabinovich.
\newblock Going deeper with convolutions.
\newblock {\em CoRR}, abs/1409.4842, 2014.

\bibitem{8500456}
Álvaro Sáez, Luis~M. Bergasa, Eduardo Romeral, Elena López, Rafael Barea,
  and Rafael Sanz.
\newblock Cnn-based fisheye image real-time semantic segmentation.
\newblock In {\em 2018 IEEE Intelligent Vehicles Symposium (IV)}, 2018.

\bibitem{torralba2011unbiased}
Antonio Torralba and Alexei~A Efros.
\newblock Unbiased look at dataset bias.
\newblock In {\em IEEE Conf. Comput. Vis. Pattern Recog.}, 2011.

\bibitem{vaswani2017attention}
Ashish Vaswani, Noam Shazeer, Niki Parmar, Jakob Uszkoreit, Llion Jones,
  Aidan~N Gomez, {\L}ukasz Kaiser, and Illia Polosukhin.
\newblock Attention is all you need.
\newblock In {\em Adv. Neural Inform. Process. Syst.}, 2017.

\bibitem{xia2022vision}
Zhuofan Xia, Xuran Pan, Shiji Song, Li~Erran Li, and Gao Huang.
\newblock Vision transformer with deformable attention.
\newblock In {\em IEEE Conf. Comput. Vis. Pattern Recog.}, 2022.

\bibitem{xue2019learning}
Z. Xue, N. Xue, G. Xia, and W. Shen.
\newblock Learning to calibrate straight lines for fisheye image rectification.
\newblock In {\em IEEE Conf. Comput. Vis. Pattern Recog.}, 2019.

\bibitem{Ye2020UniversalSS}
Yaozu Ye, Kailun Yang, Kaite Xiang, Juan Wang, and Kaiwei Wang.
\newblock Universal semantic segmentation for fisheye urban driving images.
\newblock {\em IEEE Int. Conf. Syst. Man Cyber.}, pages 648--655, 2020.

\bibitem{yin2018fisheyerectnet}
Xiaoqing Yin, Xinchao Wang, Jun Yu, Maojun Zhang, Pascal Fua, and Dacheng Tao.
\newblock Fisheyerecnet: A multi-context collaborative deep network for fisheye
  image rectification.
\newblock In {\em Eur. Conf. Comput. Vis.}, 2018.

\bibitem{ying2004consider}
Xianghua Ying and Zhanyi Hu.
\newblock Can we consider central catadioptric cameras and fisheye cameras
  within a unified imaging model.
\newblock In {\em Eur. Conf. Comput. Vis.}, 2004.

\bibitem{yogamani2019woodscapes}
Senthil Yogamani, Ciaran Hughes, Jonathan Horgan, Ganesh Sistu, Sumanth
  Chennupati, Michal Uricar, Stefan Milz, Martin Simon, Karl Amende, Christian
  Witt, Hazem Rashed, Sanjaya Nayak, Saquib Mansoor, Padraig Varley, Xavier
  Perrotton, Derek Odea, and Patrick Pérez.
\newblock Woodscape: A multi-task, multi-camera fisheye dataset for autonomous
  driving.
\newblock In {\em Int. Conf. Comput. Vis.}, 2019.

\bibitem{yun2022panoramic}
Heeseung Yun, Sehun Lee, and Gunhee Kim.
\newblock Panoramic vision transformer for saliency detection in \ang{360}
  videos.
\newblock In {\em Eur. Conf. Comput. Vis.}, 2022.

\bibitem{zhang2022bending}
Jiaming Zhang, Kailun Yang, Chaoxiang Ma, Simon Rei{\ss}, Kunyu Peng, and
  Rainer Stiefelhagen.
\newblock Bending reality: Distortion-aware transformers for adapting to
  panoramic semantic segmentation.
\newblock In {\em IEEE Conf. Comput. Vis. Pattern Recog.}, 2022.

\bibitem{zhang2015line}
Mi Zhang, Jian Yao, Menghan Xia, Kai Li, Yi Zhang, and Yaping Liu.
\newblock Line-based multi-label energy optimization for fisheye image
  rectification and calibration.
\newblock In {\em IEEE Conf. Comput. Vis. Pattern Recog.}, 2015.

\bibitem{zhou2021deepvit}
Daquan Zhou, Bingyi Kang, Xiaojie Jin, Linjie Yang, Xiaochen Lian, Qibin Hou,
  and Jiashi Feng.
\newblock Deepvit: Towards deeper vision transformer.
\newblock {\em arXiv preprint arXiv:2103.11886}, 2021.

\bibitem{zhu2019deformable}
Xizhou Zhu, Han Hu, Stephen Lin, and Jifeng Dai.
\newblock Deformable convnets v2: More deformable, better results.
\newblock In {\em IEEE Conf. Comput. Vis. Pattern Recog.}, 2019.

\bibitem{zioulis2018omnidepth}
Nikolaos Zioulis, Antonis Karakottas, Dimitrios Zarpalas, and Petros Daras.
\newblock Omnidepth: Dense depth estimation for indoors spherical panoramas.
\newblock In {\em Eur. Conf. Comput. Vis.}, 2018.

\end{thebibliography}
}



\end{document}